\documentclass[11pt,a4paper]{article}
\usepackage[hyperref]{emnlp2020}
\usepackage{times}
\usepackage{latexsym}

\usepackage{booktabs}
\usepackage{multirow}
\usepackage{amsmath}
\usepackage{amssymb}
\usepackage{colortbl}
\usepackage{xspace}
\usepackage{pifont}
\usepackage{graphicx}
\usepackage{textcomp}
\usepackage{url}

\newcommand{\inc}[2]{\cellcolor{cyan!#1}#2}
\newcommand{\cmark}{\ding{51}}
\newcommand{\xmark}{\ding{55}}

\usepackage{microtype}

\aclfinalcopy 
\def\aclpaperid{2615} 

\title{Calibration of Pre-trained Transformers}

\author{Shrey Desai \and Greg Durrett \\
  Department of Computer Science\\
  The University of Texas at Austin \\
  \texttt{shreydesai@utexas.edu\quad gdurrett@cs.utexas.edu}}

\date{}

\begin{document}
\maketitle
\begin{abstract}
Pre-trained Transformers are now ubiquitous in natural language processing, but despite their high end-task performance, little is known empirically about whether they are calibrated. Specifically, do these models' posterior probabilities provide an accurate empirical measure of how likely the model is to be correct on a given example? We focus on BERT \cite{devlin-etal-2019-bert} and RoBERTa \cite{liu2019roberta} in this work, and analyze their calibration across three tasks: natural language inference, paraphrase detection, and commonsense reasoning. For each task, we consider in-domain as well as challenging out-of-domain settings, where models face more examples they \emph{should} be uncertain about. We show that: (1) when used out-of-the-box, pre-trained models are calibrated in-domain, and compared to baselines, their calibration error out-of-domain can be as much as 3.5$\times$ lower; (2) temperature scaling is effective at further reducing calibration error in-domain, and using label smoothing to deliberately increase empirical uncertainty helps calibrate posteriors out-of-domain.\footnote{Code and datasets available at \url{https://github.com/shreydesai/calibration}}
\end{abstract}

\section{Introduction}

Neural networks have seen wide adoption but are frequently criticized for being black boxes, offering little insight as to why predictions are made \cite{benitez-are-1997,dayhoff-artificial-2001,castelvecchi-can-2016} and making it difficult to diagnose errors at test-time. These properties are particularly exhibited by pre-trained Transformer models \cite{devlin-etal-2019-bert,liu2019roberta,yang-2019-xlnet}, which dominate benchmark tasks like SuperGLUE \cite{wang-2019-superglue}, but use a large number of self-attention heads across many layers in a way that is difficult to unpack \cite{clark-etal-2019-bert,kovaleva-etal-2019-revealing}. One step towards understanding whether these models can be trusted is by analyzing whether they are \emph{calibrated} \cite{raftery-using-2005,jiang-calibrating-2012,kendall-what-2017}: how aligned their posterior probabilities are with empirical likelihoods \cite{brier-verification-1950,guo-on-2017}. If a model assigns 70\% probability to an event, the event should occur 70\% of the time if the model is calibrated. Although the model's mechanism itself may be uninterpretable, a calibrated model at least gives us a signal that it ``knows what it doesn't know,'' which can make these models easier to deploy in practice \cite{jiang-calibrating-2012}.

In this work, we evaluate the calibration of two pre-trained models, BERT \cite{devlin-etal-2019-bert} and RoBERTa \cite{liu2019roberta}, on three tasks: natural language inference \cite{bowman-etal-2015-large}, paraphrase detection \cite{iyer-2017-quora}, and commonsense reasoning \cite{zellers-etal-2018-swag}. These tasks represent standard evaluation settings for pre-trained models, and critically, challenging out-of-domain test datasets are available for each. Such test data allows us to measure calibration in more realistic settings where samples stem from a dissimilar input distribution, which is exactly the scenario where we hope a well-calibrated model would avoid making confident yet incorrect predictions.

Our experiments yield several key results. First, even when used out-of-the-box, pre-trained models are calibrated in-domain. In out-of-domain settings, where non-pre-trained models like ESIM \cite{chen-etal-2017-enhanced} are overconfident, we find that pre-trained models are significantly better calibrated. Second, we show that temperature scaling \cite{guo-on-2017}, multiplying non-normalized logits by a single scalar hyperparameter, is widely effective at improving in-domain calibration. Finally, we show that regularizing the model to be less certain during training can beneficially smooth probabilities, improving out-of-domain calibration.

\section{Related Work}

Calibration has been well-studied in statistical machine learning, including applications in forecasting \cite{brier-verification-1950,raftery-using-2005,gneiting2007probabilistic,palmer2008toward}, medicine \cite{yang-nurses-2010,jiang-calibrating-2012}, and computer vision \cite{kendall-what-2017,guo-on-2017,lee2018training}. Past work in natural language processing has studied calibration in the non-neural \cite{nguyen-oconnor-2015-posterior} and neural \cite{kumar-2019-calibration} settings across several tasks. However, past work has not analyzed large-scale pre-trained models, and we additionally analyze out-of-domain settings, whereas past work largely focuses on in-domain calibration \cite{nguyen-oconnor-2015-posterior,guo-on-2017}.

Another way of hardening models against out-of-domain data is to be able to explicitly detect these examples, which has been studied previously \cite{hendrycks2016a,liang2018enhancing,lee2018training}. However, this assumes a discrete notion of domain; calibration is a more general paradigm and gracefully handles settings where domains are less quantized.

\section{Posterior Calibration}
\label{sec:posterior-calibration}

A model is calibrated if the confidence estimates of its predictions are aligned with empirical likelihoods. For example, if we take 100 samples where a model's prediction receives posterior probability 0.7, the model should get 70 of the samples correct. Formally, calibration is expressed as a joint distribution $P(Q,Y)$ over confidences $Q \in \mathbb{R}$ and labels $Y \in \mathcal{Y}$, where perfect calibration is achieved when $P(Y=y|Q=q)=q$. This probability can be empirically approximated by binning predictions into $k$ disjoint, equally-sized bins, each consisting of $b_k$ predictions. Following previous work in measuring calibration \cite{guo-on-2017}, we use \textit{expected calibration error} (ECE), which is a weighted average of the difference between each bin's accuracy and confidence: $\sum_k \frac{b_k}{n} |\mathrm{acc}(k) - \mathrm{conf}(k)|$. For the experiments in this paper, we use $k=10$.

\section{Experiments}
\label{sec:experiments}

\begin{table}[ht!]
\centering
\small
\begin{tabular}{lccc}
\toprule
Model & Parameters & Architecture & Pre-trained \\
\midrule
DA & 382K & LSTM & \xmark \\
ESIM & 4M & Bi-LSTM & \xmark \\
BERT & 110M & Transformer & \cmark \\
RoBERTa & 110M & Transformer & \cmark \\
\bottomrule
\end{tabular}
\caption{Models in this work. Decomposable Attention (DA) \cite{parikh-etal-2016-decomposable} and Enhanced Sequential Inference Model (ESIM) \cite{chen-etal-2017-enhanced} use LSTMs and attention on top of GloVe embeddings \cite{pennington2014glove} to model pairwise semantic similarities. In contrast, BERT \cite{devlin-etal-2019-bert} and RoBERTa \cite{liu2019roberta} are large-scale, pre-trained language models with stacked, general purpose Transformer \cite{vaswani2017attention} layers.}
\label{table:models}
\end{table}

\subsection{Tasks and Datasets}
\label{sec:tasks-and-datasets}

We perform evaluations on three language understanding tasks: natural language inference, paraphrase detection, and commonsense reasoning. Significant past work has studied cross-domain robustness using sentiment analysis \cite{chen-etal-2018-adversarial,peng-etal-2018-cross,miller-2019-simplified,desai-etal-2019-evaluating}. However, we explicitly elect to use tasks where out-of-domain performance is substantially lower and challenging domain shifts are exhibited. Below, we describe our in-domain and out-of-domain datasets.\footnote{Dataset splits are detailed in Appendix \ref{sec:splits}. Furthermore, out-of-domain datasets are strictly used for evaluating the generalization of in-domain models, so the training split is unused.} For all datasets, we split the development set in half to obtain a held-out, non-blind test set.

\paragraph{Natural Language Inference.} The Stanford Natural Language Inference (SNLI) corpus is a large-scale entailment dataset where the task is to determine whether a hypothesis is entailed, contradicted by, or neutral with respect to a premise \cite{bowman-etal-2015-large}. Multi-Genre Natural Language Inference (MNLI) \cite{williams-etal-2018-broad} contains similar entailment data across several domains, which we can use as unseen test domains.

\paragraph{Paraphrase Detection.} Quora Question Pairs (QQP) contains sentence pairs from Quora that are semantically equivalent \cite{iyer-2017-quora}. Our out-of-domain setting is TwitterPPDB (TPPDB), which contains sentence pairs from Twitter where tweets are considered paraphrases if they have shared URLs \cite{lan-etal-2017-continuously}. 

\paragraph{Commonsense Reasoning.} Situations With Adversarial Generations (SWAG) is a grounded commonsense reasoning task where models must select the most plausible continuation of a sentence among four candidates \cite{zellers-etal-2018-swag}. HellaSWAG (HSWAG), an adversarial out-of-domain dataset, serves as a more challenging benchmark for pre-trained models \cite{zellers-etal-2019-hellaswag}; it is distributionally different in that its examples exploit statistical biases in pre-trained models.

\subsection{Systems for Comparison}

Table \ref{table:models} shows a breakdown of the models used in our experiments. We use the same set of hyperparameters across all tasks. For pre-trained models, we omit hyperparameters that induce brittleness during fine-tuning, e.g., employing a decaying learning rate schedule with linear warmup \cite{sun2019finetune,lan-albert-2020}. Detailed information on optimization is available in Appendix~\ref{sec:training-and-optimization}.

\begin{table}[t!]
\centering
\small
\begin{tabular}{lrrrr}
\toprule
\multirow{2}{*}{Model} & \multicolumn{2}{c}{Accuracy} & \multicolumn{2}{c}{ECE} \\
\cmidrule(lr){2-3} \cmidrule(lr){4-5}
 & ID & OD & ID & OD \\
\midrule
\multicolumn{5}{l}{\textbf{Task: SNLI/MNLI}} \\
\midrule
DA & 84.63 & 57.12 & \textbf{1.02} & 8.79 \\
ESIM & 88.32 & 60.91 & 1.33 & 12.78 \\
BERT & 90.04 & 73.52 & 2.54 & 7.03 \\
RoBERTa & \textbf{91.23} & \textbf{78.79} & 1.93 & \textbf{3.62} \\
\midrule
\multicolumn{5}{l}{\textbf{Task: QQP/TwitterPPDB}} \\
\midrule
DA & 85.85 & 83.36 & 3.37 & 9.79 \\
ESIM & 87.75 & 84.00 & 3.65 & \textbf{8.38} \\
BERT & 90.27 & \textbf{87.63} & 2.71 & 8.51 \\
RoBERTa & \textbf{91.11} & 86.72 & \textbf{2.33} & 9.55 \\
\midrule
\multicolumn{5}{l}{\textbf{Task: SWAG/HellaSWAG}} \\
\midrule
DA & 46.80 & 32.48 & 5.98 & 40.37 \\
ESIM & 52.09 & 32.08 & 7.01 & 19.57 \\
BERT & 79.40 & 34.48 & 2.49 & 12.62 \\
RoBERTa & \textbf{82.45} & \textbf{41.68} & \textbf{1.76} & \textbf{11.93} \\
\bottomrule
\end{tabular}
\caption{Out-of-the-box calibration results for in-domain (SNLI, QQP, SWAG) and out-of-domain (MNLI, TwitterPPDB, HellaSWAG) datasets using the models described in Table \ref{table:models}. We report accuracy and expected calibration error (ECE), both averaged across 5 fine-tuning runs with random restarts.}
\label{table:out-of-the-box-calibration}
\end{table}

\begin{figure}[t]
    \begin{center}
        \includegraphics[scale=0.45]{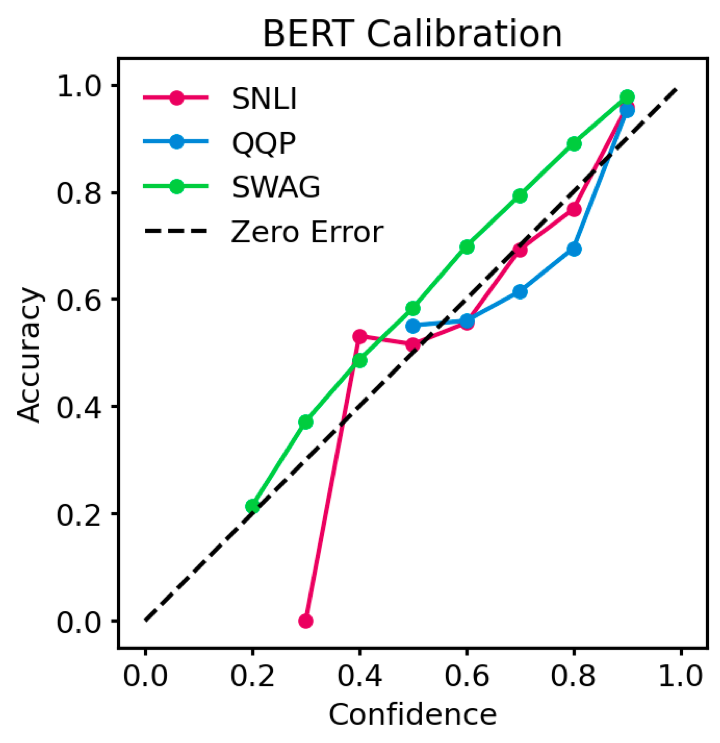}
        \includegraphics[scale=0.45]{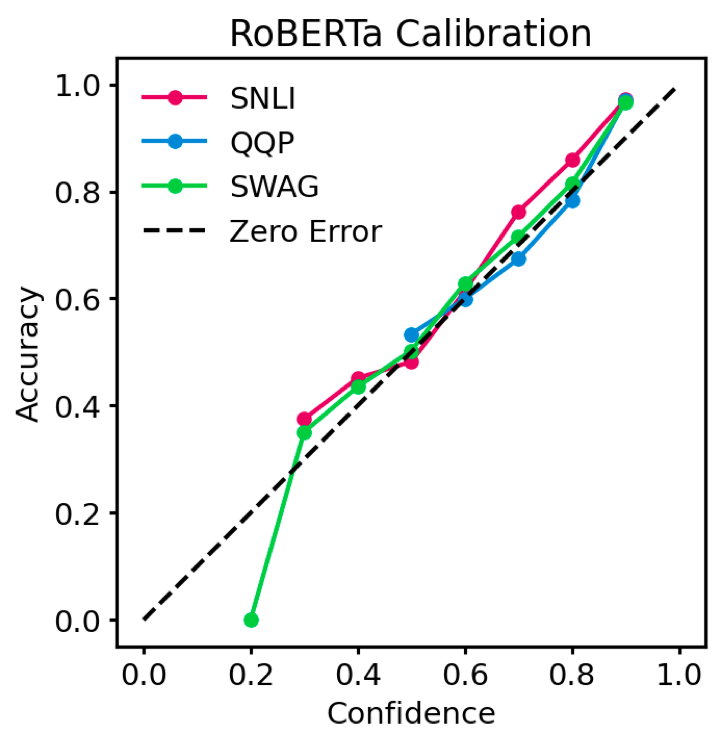}
    \end{center}
    \caption{In-domain calibration of BERT and RoBERTa when used out-of-the-box. Models are both trained and evaluated on SNLI, QQP, and SWAG, respectively. \textsc{Zero Error} depicts perfect calibration (e.g., expected calibration error = 0). Note that low-confidence buckets have zero accuracy due to a small sample count; however, as a result, these buckets do not influence the expected error as much.}
    \label{fig:reliability-out-of-the-box}
\end{figure}

\subsection{Out-of-the-box Calibration}
\label{sec:out-of-the-box-calibration}

First, we analyze ``out-of-the-box'' calibration; that is, the calibration error derived from evaluating a model on a dataset without using post-processing steps like temperature scaling \cite{guo-on-2017}. For each task, we train the model on the in-domain training set, and then evaluate its performance on the in-domain and out-of-domain test sets. Quantitative results are shown in Table \ref{table:out-of-the-box-calibration}. In addition, we plot reliability diagrams \cite{nguyen-oconnor-2015-posterior,guo-on-2017} in Figure \ref{fig:reliability-out-of-the-box}, which visualize the alignment between posterior probabilities (confidence) and empirical outcomes (accuracy), where a perfectly calibrated model has $\textrm{conf}(k) = \textrm{acc}(k)$ for each bucket of real-valued predictions $k$. We remark on a few observed phenomena below:

\paragraph{Non-pre-trained models exhibit an inverse relationship between complexity and calibration.} Simpler models, such as DA, achieve competitive in-domain ECE on SNLI (1.02) and QQP (3.37), and are notably better than pre-trained models on SNLI in this regard. However, the more complex ESIM, both in number of parameters and architecture, sees increased in-domain ECE despite having higher accuracy on all tasks.

\paragraph{However, pre-trained models are generally more accurate and calibrated.} Rather surprisingly, pre-trained models do not show characteristics of the aforementioned inverse relationship, despite having significantly more parameters. On SNLI, RoBERTa achieves an ECE in the ballpark of DA and ESIM, but on QQP and SWAG, both BERT and RoBERTa consistently achieve higher accuracies and lower ECEs. Pre-trained models are especially strong out-of-domain, where on HellaSWAG in particular, RoBERTa reduces ECE by a factor of 3.4 compared to DA.

\paragraph{Using RoBERTa always improves in-domain calibration over BERT.} In addition to obtaining better task performance than BERT, RoBERTa consistently achieves lower in-domain ECE. Even out-of-domain, RoBERTa outperforms BERT in all but one setting (TwitterPPDB). Nonetheless, our results show that representations induced by robust pre-training (e.g., using a larger corpus, more training steps, dynamic masking) \cite{liu2019roberta} lead to more calibrated posteriors. Whether other changes to pre-training \cite{yang-2019-xlnet,lan-albert-2020,clark2020electra} lead to further improvements is an open question.

\subsection{Post-hoc Calibration}
\label{sec:posthoc-calibration}

\begin{table*}[ht]
\centering
\small
\setlength{\tabcolsep}{4pt}
\begin{tabular}{lcccccccccccc}
\toprule
\multirow{4}{*}{Method} & \multicolumn{6}{c}{In-Domain} & \multicolumn{6}{c}{Out-of-Domain} \\
 \cmidrule(lr){2-7} \cmidrule(lr){8-13}
 & \multicolumn{2}{c}{SNLI} & \multicolumn{2}{c}{QQP} & \multicolumn{2}{c}{SWAG} & \multicolumn{2}{c}{MNLI} & \multicolumn{2}{c}{TPPDB} & \multicolumn{2}{c}{HSWAG} \\
 \cmidrule(lr){2-3} \cmidrule(lr){4-5} \cmidrule(lr){6-7} \cmidrule(lr){8-9} \cmidrule(lr){10-11} \cmidrule(lr){12-13}
 & MLE & LS & MLE & LS & MLE & LS & MLE & LS & MLE & LS & MLE & LS \\
\midrule
\multicolumn{13}{l}{\textbf{Model: BERT}} \\
\midrule
Out-of-the-box & \inc{74.60}{2.54} & \inc{28.80}{7.12} & \inc{72.90}{2.71} & \inc{36.70}{6.33} & \inc{75.10}{2.49} & \inc{0.00}{10.01} & \inc{29.70}{7.03} & \inc{62.60}{3.74} & \inc{14.90}{8.51} & \inc{37.00}{6.30} & \inc{0.00}{12.62} & \inc{42.70}{5.73} \\
Temperature scaled & \inc{88.60}{1.14} & \inc{16.30}{8.37} & \inc{90.30}{0.97} & \inc{18.40}{8.16} & \inc{91.50}{0.85} & \inc{0.00}{10.89} & \inc{63.90}{3.61} & \inc{59.50}{4.05} & \inc{28.50}{7.15} & \inc{42.20}{5.78} & \inc{0.00}{12.83} & \inc{46.60}{5.34} \\
\midrule
\multicolumn{13}{l}{\textbf{Model: RoBERTa}} \\
\midrule
Out-of-the-box & \inc{80.70}{1.93} & \inc{36.20}{6.38} & \inc{76.70}{2.33} & \inc{38.90}{6.11} & \inc{82.40}{1.76} & \inc{11.90}{8.81} & \inc{63.80}{3.62} & \inc{55.00}{4.50} & \inc{4.50}{9.55} & \inc{10.90}{8.91} & \inc{0.00}{11.93} & \inc{78.60}{2.14} \\
Temperature scaled & \inc{91.60}{0.84} & \inc{13.00}{8.70} & \inc{91.20}{0.88} & \inc{13.10}{8.69} & \inc{92.40}{0.76} & \inc{0.00}{11.40} & \inc{85.40}{1.46} & \inc{40.70}{5.93} & \inc{21.40}{7.86} & \inc{46.90}{5.31} & \inc{0.00}{11.22} & \inc{77.70}{2.23} \\
\bottomrule
\end{tabular}
\caption{Post-hoc calibration results for BERT and RoBERTa on in-domain (SNLI, QQP, SWAG) and out-of-domain (MNLI, TwitterPPDB, HellaSWAG) datasets. Models are trained with maximum likelihood estimation (MLE) or label smoothing (LS), then their logits are post-processed using temperature scaling (\S\ref{sec:posthoc-calibration}). We report expected calibration error (ECE) averaged across 5 runs with random restarts. Darker colors imply lower ECE.}
\label{table:posthoc-calibration}
\end{table*}

There are a number of techniques that can be applied to correct a model's calibration post-hoc. Using our in-domain development set, we can, for example, post-process model probabilities via temperature scaling \cite{guo-on-2017}, where a scalar temperature hyperparameter $T$ divides non-normalized logits before the softmax operation. As $T \rightarrow 0$, the distribution's mode receives all the probability mass, while as $T \rightarrow \infty$, the probabilities become uniform.

Furthermore, we experiment with models trained in-domain with label smoothing (LS) \cite{miller1996a,pereyra2017regularizing} as opposed to conventional maximum likelihood estimation (MLE). By nature, MLE encourages models to sharpen the posterior distribution around the gold label, leading to confidence which is typically unwarranted in out-of-domain settings. Label smoothing presents one solution to overconfidence by maintaining uncertainty over the label space during training: we minimize the KL divergence with the distribution placing a $1-\alpha$ fraction of probability mass on the gold label and $\frac{\alpha}{|\mathcal{Y}|-1}$ fraction of mass on each other label, where $\alpha \in (0, 1)$ is a hyperparameter.\footnote{For example, the one-hot target [1, 0, 0] is transformed into [0.9, 0.05, 0.05] when $\alpha=0.1$.} This re-formulated learning objective does not require changing the model architecture.

\begin{figure}[t]
    \begin{center}
        \includegraphics[scale=0.45]{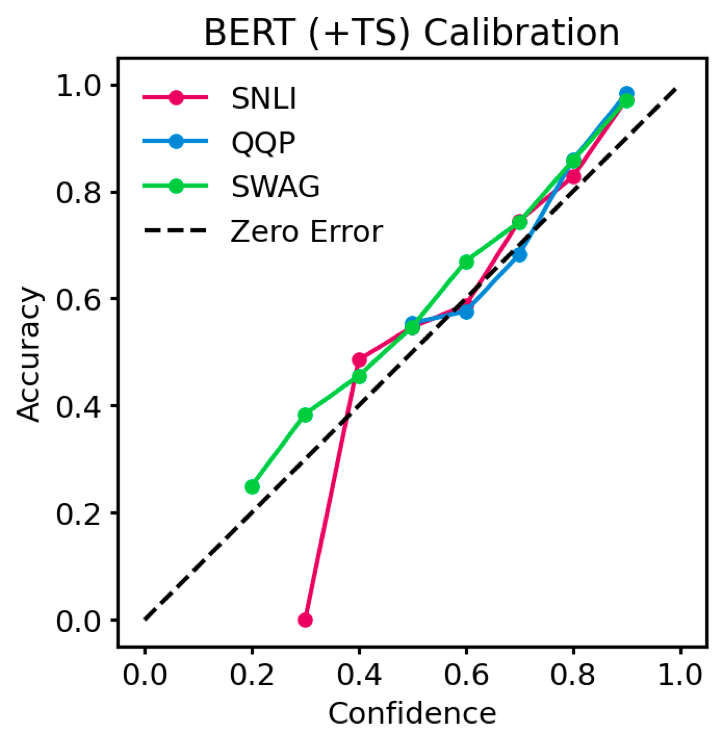}
        \includegraphics[scale=0.45]{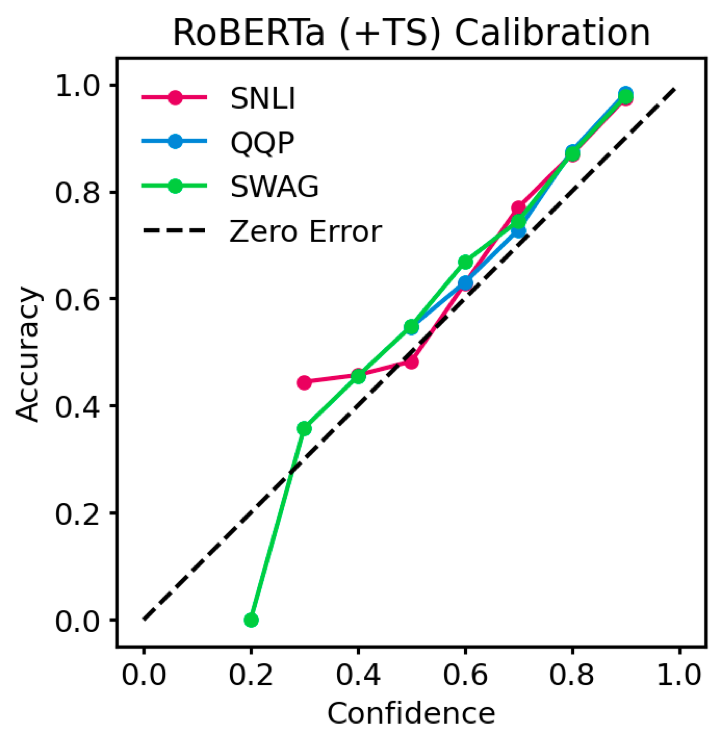}
    \end{center}
    \caption{In-domain calibration of BERT and RoBERTa with temperature scaling (TS). Both temperature-scaled models are much better calibrated than when used out-of-the-box, with BERT especially showing a large degree of improvement.}
    \label{fig:reliability-post-processed}
\end{figure}

For each task, we train the model with either MLE or LS ($\alpha=0.1$) using the in-domain training set, use the in-domain development set to learn an optimal temperature $T$, and then evaluate the model (scaled with $T$) on the in-domain and out-of-domain test sets. From Table \ref{table:posthoc-calibration} and Figure \ref{fig:reliability-post-processed}, we draw the following conclusions:

\paragraph{MLE models with temperature scaling achieve low in-domain calibration error.} MLE models are always better than LS models in-domain, which suggests incorporating uncertainty when in-domain samples are available is not an effective regularization scheme. Even when using a small smoothing value (0.1), LS models do not achieve nearly as good out-of-the-box results as MLE models, and temperature scaling hurts LS in many cases. By contrast, RoBERTa with temperature-scaled MLE achieves ECE values from 0.7-0.8, implying that MLE training yields scores that are fundamentally good but just need some minor rescaling.

\begin{figure}[t]
    \centering
    \includegraphics[scale=0.45]{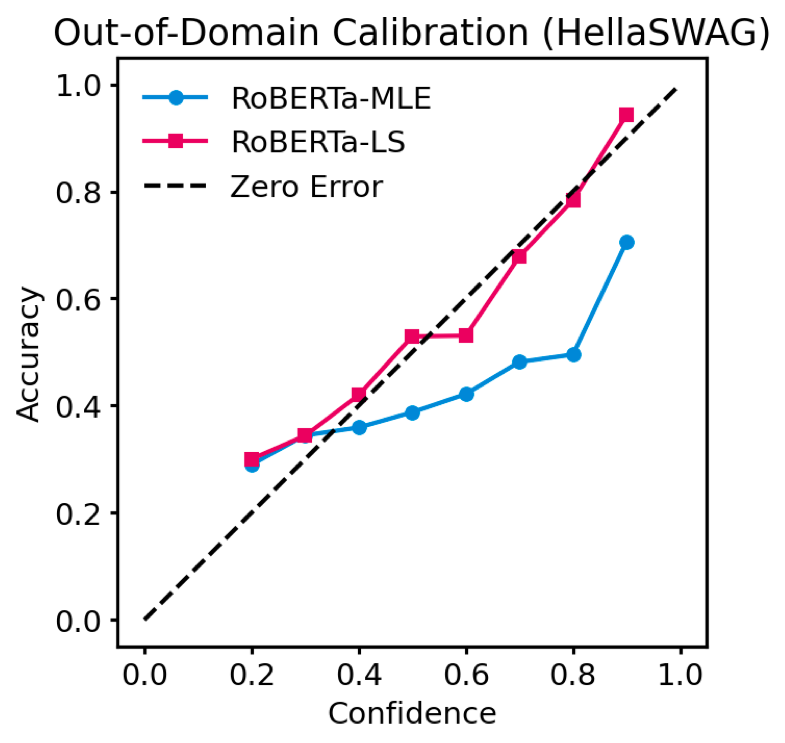}
    \caption{Out-of-domain calibration of RoBERTa fine-tuned on SWAG with different learning objectives and used out-of-the-box on HellaSWAG. Without seeing HellaSWAG samples during fine-tuning, RoBERTa-LS achieves significantly lower calibration error than RoBERTa-MLE.}
    \label{fig:reliability-ood}
\end{figure}

\paragraph{However, out-of-domain, label smoothing is generally more effective.} In most cases, MLE models do not perform well on out-of-domain datasets, with ECEs ranging from 8-12. However, LS models are forced to distribute probability mass across classes, and as a result, achieve significantly lower ECEs on average. We note that LS is particularly effective when the distribution shift is strong. On the adversarial HellaSWAG, for example, RoBERTa-LS obtains a factor of 5.8 less ECE than RoBERTa-MLE. This phenomenon is visually depicted in Figure \ref{fig:reliability-ood} where we see RoBERTa-LS is significantly closer to the identity function despite being used out-of-the-box.

\begin{table}[t]
\centering
\small
\setlength{\tabcolsep}{2.5pt}
\begin{tabular}{lrrrrrr}
\toprule
\multirow{2}{*}{Model} & \multicolumn{3}{c}{In-Domain} & \multicolumn{3}{c}{Out-of-Domain} \\
 \cmidrule(lr){2-4} \cmidrule(lr){5-7}
 & SNLI & QQP & SWAG & MNLI & TPPDB & HSWAG \\
\midrule
BERT & 1.20 & 1.34 & 0.99 & 1.41 & 2.91 & 3.61 \\
RoBERTa & 1.16 & 1.39 & 1.10 & 1.25 & 2.79 & 2.77 \\
\bottomrule
\end{tabular}
\caption{Learned temperature scaling values for BERT and RoBERTa on in-domain (SNLI, QQP, SWAG) and out-of-domain (MNLI, TwitterPPDB, HellaSWAG) datasets. Values are obtained by line search with a granularity of 0.01. Evaluations are very fast as they only require rescaling cached logits.}
\label{table:temperature-scaling}
\end{table}

\paragraph{Optimal temperature scaling values are bounded within a small interval.} Table \ref{table:temperature-scaling} reports the learned temperature values for BERT-MLE and RoBERTa-MLE. For in-domain tasks, the optimal temperature values are generally in the range 1-1.4. Interestingly, out-of-domain, TwitterPPDB and HellaSWAG require larger temperature values than MNLI, which suggests the degree of distribution shift and magnitude of $T$ may be closely related.

\section{Conclusion}

Posterior calibration is one lens to understand the trustworthiness of model confidence scores. In this work, we examine the calibration of pre-trained Transformers in both in-domain and out-of-domain settings. Results show BERT and RoBERTa coupled with temperature scaling achieve low ECEs in-domain, and when trained with label smoothing, are also competitive out-of-domain.

\section*{Acknowledgments}

This work was partially supported by NSF Grant IIS-1814522 and a gift from Arm. The authors acknowledge a DURIP equipment grant to UT Austin that provided computational resources to conduct this research. Additionally, we thank R. Thomas McCoy for answering questions about DA and ESIM.

\bibliography{emnlp2020}

\begin{thebibliography}{44}
\expandafter\ifx\csname natexlab\endcsname\relax\def\natexlab#1{#1}\fi

\bibitem[{Benítez et~al.(1997)Benítez, Castro, and
  Requena}]{benitez-are-1997}
Jose~M. Benítez, Juan~Luis Castro, and Ignacio Requena. 1997.
\newblock {Are Artificial Neural Networks Black Boxes?}
\newblock \emph{IEEE Transactions on Neural Networks and Learning Systems}.

\bibitem[{Bowman et~al.(2015)Bowman, Angeli, Potts, and
  Manning}]{bowman-etal-2015-large}
Samuel~R. Bowman, Gabor Angeli, Christopher Potts, and Christopher~D. Manning.
  2015.
\newblock {A Large Annotated Corpus for Learning Natural Language Inference}.
\newblock In \emph{Proceedings of the Conference on Empirical Methods in
  Natural Language Processing (EMNLP)}.

\bibitem[{Brier(1950)}]{brier-verification-1950}
Glenn~W. Brier. 1950.
\newblock {Verification of Forecasts Expressed in Terms of Probability}.
\newblock \emph{Monthly Weather Review}.

\bibitem[{Castelvecchi(2016)}]{castelvecchi-can-2016}
Davide Castelvecchi. 2016.
\newblock {Can We Open the Black Box of AI?}
\newblock \emph{Nature News}.

\bibitem[{Chen et~al.(2017)Chen, Zhu, Ling, Wei, Jiang, and
  Inkpen}]{chen-etal-2017-enhanced}
Qian Chen, Xiaodan Zhu, Zhen-Hua Ling, Si~Wei, Hui Jiang, and Diana Inkpen.
  2017.
\newblock {Enhanced LSTM for Natural Language Inference}.
\newblock In \emph{Proceedings of the Annual Meeting of the Association for
  Computational Linguistics (ACL)}.

\bibitem[{Chen et~al.(2018)Chen, Sun, Athiwaratkun, Cardie, and
  Weinberger}]{chen-etal-2018-adversarial}
Xilun Chen, Yu~Sun, Ben Athiwaratkun, Claire Cardie, and Kilian Weinberger.
  2018.
\newblock {Adversarial Deep Averaging Networks for Cross-Lingual Sentiment
  Classification}.
\newblock \emph{Transactions of the Association for Computational Linguistics
  (TACL)}.

\bibitem[{Clark et~al.(2019)Clark, Khandelwal, Levy, and
  Manning}]{clark-etal-2019-bert}
Kevin Clark, Urvashi Khandelwal, Omer Levy, and Christopher~D. Manning. 2019.
\newblock {What Does BERT Look at? An Analysis of BERT's Attention}.
\newblock In \emph{Proceedings of the Workshop on BlackboxNLP}.

\bibitem[{Clark et~al.(2020)Clark, Luong, Le, and Manning}]{clark2020electra}
Kevin Clark, Minh-Thang Luong, Quoc~V. Le, and Christopher~D. Manning. 2020.
\newblock {ELECTRA: Pre-training Text Encoders as Discriminators Rather Than
  Generators}.
\newblock In \emph{Proceedings of the International Conference on Learning
  Representations (ICLR)}.

\bibitem[{Dayhoff and DeLeo(2001)}]{dayhoff-artificial-2001}
Judith~E. Dayhoff and James~M. DeLeo. 2001.
\newblock {Artificial Neural Networks: Opening the Black Box}.
\newblock \emph{Cancer: Interdisciplinary International Journal of the American
  Cancer Society}.

\bibitem[{Desai et~al.(2019)Desai, Zhan, and Aly}]{desai-etal-2019-evaluating}
Shrey Desai, Hongyuan Zhan, and Ahmed Aly. 2019.
\newblock {Evaluating Lottery Tickets Under Distributional Shifts}.
\newblock In \emph{Proceedings of the Workshop on Deep Learning Approaches for
  Low-Resource NLP (DeepLo)}.

\bibitem[{Devlin et~al.(2019)Devlin, Chang, Lee, and
  Toutanova}]{devlin-etal-2019-bert}
Jacob Devlin, Ming-Wei Chang, Kenton Lee, and Kristina Toutanova. 2019.
\newblock {BERT: Pre-training of Deep Bidirectional Transformers for Language
  Understanding}.
\newblock In \emph{Proceedings of the Conference of the North American Chapter
  of the Association for Computational Linguistics (NAACL)}.

\bibitem[{Gardner et~al.(2018)Gardner, Grus, Neumann, Tafjord, Dasigi, Liu,
  Peters, Schmitz, and Zettlemoyer}]{gardner-etal-2018-allennlp}
Matt Gardner, Joel Grus, Mark Neumann, Oyvind Tafjord, Pradeep Dasigi,
  Nelson~F. Liu, Matthew Peters, Michael Schmitz, and Luke Zettlemoyer. 2018.
\newblock {AllenNLP: A Deep Semantic Natural Language Processing Platform}.
\newblock In \emph{Proceedings of the Workshop for NLP Open Source Software
  (NLP-OSS)}.

\bibitem[{Gneiting et~al.(2007)Gneiting, Balabdaoui, and
  Raftery}]{gneiting2007probabilistic}
Tilmann Gneiting, Fadoua Balabdaoui, and Adrian~E. Raftery. 2007.
\newblock {Probabilistic Forecasts, Calibration and Sharpness}.
\newblock \emph{Journal of the Royal Statistical Society: Series B (Statistical
  Methodology)}.

\bibitem[{Guo et~al.(2017)Guo, Pleiss, Sun, and Weinberger}]{guo-on-2017}
Chuan Guo, Geoff Pleiss, Yu~Sun, and Kilian~Q. Weinberger. 2017.
\newblock {On Calibration of Modern Neural Networks}.
\newblock In \emph{Proceedings of the International Conference on Machine
  Learning (ICML)}.

\bibitem[{Hendrycks and Gimpel(2016)}]{hendrycks2016a}
Dan Hendrycks and Kevin Gimpel. 2016.
\newblock {A Baseline for Detecting Misclassified and Out-of-Distribution
  Examples in Neural Networks}.
\newblock In \emph{Proceedings of the International Conference on Learning
  Representations (ICLR)}.

\bibitem[{Iyer et~al.(2017)Iyer, Dandekar, and Csernai}]{iyer-2017-quora}
Shankar Iyer, Nikhil Dandekar, and Kornél Csernai. 2017.
\newblock {Quora Question Pairs}.

\bibitem[{Jiang et~al.(2012)Jiang, Osl, Kim, and
  Ohno-Machado}]{jiang-calibrating-2012}
Xiaoqian Jiang, Melanie Osl, Jihoon Kim, and Lucila Ohno-Machado. 2012.
\newblock {Calibrating Predictive Model Estimates to Support Personalized
  Medicine}.
\newblock In \emph{Journal of the American Medical Informatics Association
  (JAMIA)}.

\bibitem[{Kendall and Gal(2017)}]{kendall-what-2017}
Alex Kendall and Yarin Gal. 2017.
\newblock {What Uncertainties Do We Need in Bayesian Deep Learning for Computer
  Vision?}
\newblock In \emph{Proceedings of the Conference on Neural Information
  Processing Systems (NeurIPS)}.

\bibitem[{Kovaleva et~al.(2019)Kovaleva, Romanov, Rogers, and
  Rumshisky}]{kovaleva-etal-2019-revealing}
Olga Kovaleva, Alexey Romanov, Anna Rogers, and Anna Rumshisky. 2019.
\newblock {Revealing the Dark Secrets of BERT}.
\newblock In \emph{Proceedings of the Conference on Empirical Methods in
  Natural Language Processing (EMNLP)}.

\bibitem[{Kumar and Sarawagi(2019)}]{kumar-2019-calibration}
Aviral Kumar and Sunita Sarawagi. 2019.
\newblock {Calibration of Encoder Decoder Models for Neural Machine
  Translation}.
\newblock In \emph{Proceedings of the Workshop on Debugging Machine Learning
  Models}.

\bibitem[{Lan et~al.(2017)Lan, Qiu, He, and Xu}]{lan-etal-2017-continuously}
Wuwei Lan, Siyu Qiu, Hua He, and Wei Xu. 2017.
\newblock {A Continuously Growing Dataset of Sentential Paraphrases}.
\newblock In \emph{Proceedings of the Conference on Empirical Methods in
  Natural Language Processing (EMNLP)}.

\bibitem[{Lan et~al.(2020)Lan, Chen, Goodman, Gimpel, Sharma, and
  Soricut}]{lan-albert-2020}
Zhenzhong Lan, Mingda Chen, Sebastian Goodman, Kevin Gimpel, Piyush Sharma, and
  Radu Soricut. 2020.
\newblock {ALBERT: A Lite BERT for Self-supervised Learning of Language
  Representations}.
\newblock In \emph{Proceedings of the International Conference on Learning
  Representations (ICLR)}.

\bibitem[{Lee et~al.(2018)Lee, Lee, Lee, and Shin}]{lee2018training}
Kimin Lee, Honglak Lee, Kibok Lee, and Jinwoo Shin. 2018.
\newblock {Training Confidence-calibrated Classifiers for Detecting
  Out-of-Distribution Samples}.
\newblock In \emph{Proceedings of the International Conference on Learning
  Representations (ICLR)}.

\bibitem[{Liang et~al.(2018)Liang, Li, and Srikant}]{liang2018enhancing}
Shiyu Liang, Yixuan Li, and R.~Srikant. 2018.
\newblock {Enhancing the Reliability of Out-of-distribution Image Detection in
  Neural Networks}.
\newblock In \emph{Proceedings of the International Conference on Learning
  Representations (ICLR)}.

\bibitem[{Liu et~al.(2019)Liu, Ott, Goyal, Du, Joshi, Chen, Levy, Lewis,
  Zettlemoyer, and Stoyanov}]{liu2019roberta}
Yinhan Liu, Myle Ott, Naman Goyal, Jingfei Du, Mandar Joshi, Danqi Chen, Omer
  Levy, Mike Lewis, Luke Zettlemoyer, and Veselin Stoyanov. 2019.
\newblock {RoBERTa: A Robustly Optimized BERT Pretraining Approach}.
\newblock \emph{arXiv preprint arXiv:1907.11692}.

\bibitem[{Loshchilov and Hutter(2019)}]{loshchilov2018decoupled}
Ilya Loshchilov and Frank Hutter. 2019.
\newblock {Decoupled Weight Decay Regularization}.
\newblock In \emph{Proceedings of the International Conference on Learning
  Representations (ICLR)}.

\bibitem[{Miller et~al.(1996)Miller, Rao, Rose, and Gersho}]{miller1996a}
David~J. Miller, Ajit~V. Rao, Kenneth Rose, and Allen Gersho. 1996.
\newblock {A Global Optimization Technique for Statistical Classifier Design}.
\newblock \emph{IEEE Transactions on Signal Processing}.

\bibitem[{Miller(2019)}]{miller-2019-simplified}
Timothy Miller. 2019.
\newblock {Simplified Neural Unsupervised Domain Adaptation}.
\newblock In \emph{Proceedings of the Conference of the North American Chapter
  of the Association for Computational Linguistics (NAACL)}.

\bibitem[{Nguyen and O'Connor(2015)}]{nguyen-oconnor-2015-posterior}
Khanh Nguyen and Brendan O'Connor. 2015.
\newblock {Posterior Calibration and Exploratory Analysis for Natural Language
  Processing Models}.
\newblock In \emph{Proceedings of the Conference on Empirical Methods in
  Natural Language Processing (EMNLP)}.

\bibitem[{Palmer et~al.(2008)Palmer, Doblas-Reyes, Weisheimer, and
  Rodwell}]{palmer2008toward}
Tim Palmer, Francisco Doblas-Reyes, Antje Weisheimer, and Mark Rodwell. 2008.
\newblock {Toward Seamless Prediction: Calibration of Climate Change
  Projections using Seasonal Forecasts}.
\newblock \emph{Bulletin of the American Meteorological Society}.

\bibitem[{Parikh et~al.(2016)Parikh, Täckström, Das, and
  Uszkoreit}]{parikh-etal-2016-decomposable}
Ankur Parikh, Oscar Täckström, Dipanjan Das, and Jakob Uszkoreit. 2016.
\newblock {A Decomposable Attention Model for Natural Language Inference}.
\newblock In \emph{Proceedings of the Conference on Empirical Methods in
  Natural Language Processing (EMNLP)}.

\bibitem[{Peng et~al.(2018)Peng, Zhang, gang Jiang, and
  Huang}]{peng-etal-2018-cross}
Minlong Peng, Qi~Zhang, Yu~gang Jiang, and Xuanjing Huang. 2018.
\newblock {Cross-Domain Sentiment Classification with Target Domain Specific
  Information}.
\newblock In \emph{Proceedings of the Annual Meeting of the Association for
  Computational Linguistics (ACL)}.

\bibitem[{Pennington et~al.(2014)Pennington, Socher, and
  Manning}]{pennington2014glove}
Jeffrey Pennington, Richard Socher, and Christopher~D. Manning. 2014.
\newblock {GloVe: Global Vectors for Word Representation}.
\newblock In \emph{Proceedings of the Conference on Empirical Methods in
  Natural Language Processing (EMNLP)}.

\bibitem[{Pereyra et~al.(2017)Pereyra, Tucker, Chorowski, Łukasz Kaiser, and
  Hinton}]{pereyra2017regularizing}
Gabriel Pereyra, George Tucker, Jan Chorowski, Łukasz Kaiser, and Geoffrey
  Hinton. 2017.
\newblock {Regularizing Neural Networks by Penalizing Confident Output
  Distributions}.
\newblock In \emph{Proceedings of the International Conference on Learning
  Representations (Workshop)}.

\bibitem[{Raftery et~al.(2005)Raftery, Gneiting, Balabdaoui, and
  Polakowski}]{raftery-using-2005}
Adrian~E. Raftery, Tilmann Gneiting, Fadoua Balabdaoui, and Michael Polakowski.
  2005.
\newblock {Using Bayesian Model Averaging to Calibrate Forecast Ensembles}.
\newblock \emph{Monthly Weather Review}.

\bibitem[{Sun et~al.(2019)Sun, Qiu, Xu, and Huang}]{sun2019finetune}
Chi Sun, Xipeng Qiu, Yige Xu, and Xuanjing Huang. 2019.
\newblock {How to Fine-Tune BERT for Text Classification?}
\newblock \emph{arXiv preprint arXiv:1905.05583}.

\bibitem[{Vaswani et~al.(2017)Vaswani, Shazeer, Parmar, Uszkoreit, Jones,
  Gomez, Łukasz Kaiser, and Polosukhin}]{vaswani2017attention}
Ashish Vaswani, Noam Shazeer, Niki Parmar, Jakob Uszkoreit, Llion Jones,
  Aidan~N. Gomez, Łukasz Kaiser, and Illia Polosukhin. 2017.
\newblock {Attention is All You Need}.
\newblock In \emph{Proceedings of the Conference on Neural Information
  Processing Systems (NeurIPS)}.

\bibitem[{Wang et~al.(2019)Wang, Pruksachatkun, Nangia, Singh, Michael, Hill,
  Levy, and Bowman}]{wang-2019-superglue}
Alex Wang, Yada Pruksachatkun, Nikita Nangia, Amanpreet Singh, Julian Michael,
  Felix Hill, Omer Levy, and Samuel~R. Bowman. 2019.
\newblock {SuperGLUE: A Stickier Benchmark for General-Purpose Language
  Understanding Systems}.
\newblock In \emph{Proceedings of the Conference on Neural Information
  Processing Systems (NeurIPS)}.

\bibitem[{Williams et~al.(2018)Williams, Nangia, and
  Bowman}]{williams-etal-2018-broad}
Adina Williams, Nikita Nangia, and Samuel~R. Bowman. 2018.
\newblock {A Broad-Coverage Challenge Corpus for Sentence Understanding through
  Inference}.
\newblock In \emph{Proceedings of the Conference of the North American Chapter
  of the Association for Computational Linguistics (NAACL)}.

\bibitem[{Wolf et~al.(2019)Wolf, Debut, Sanh, Chaumond, Delangue, Moi, Cistac,
  Rault, Louf, Funtowicz, and Brew}]{wolf2019huggingfaces}
Thomas Wolf, Lysandre Debut, Victor Sanh, Julien Chaumond, Clement Delangue,
  Anthony Moi, Pierric Cistac, Tim Rault, Rémi Louf, Morgan Funtowicz, and
  Jamie Brew. 2019.
\newblock {HuggingFace's Transformers: State-of-the-art Natural Language
  Processing}.
\newblock \emph{arXiv preprint arXiv:1910.03771}.

\bibitem[{Yang and Thompson(2010)}]{yang-nurses-2010}
Huiqin Yang and Carl Thompson. 2010.
\newblock {Nurses' Risk Assessment Judgements: A Confidence Calibration Study}.
\newblock \emph{Journal of Advanced Nursing}.

\bibitem[{Yang et~al.(2019)Yang, Dai, Yang, Carbonell, Salakhutdinov, and
  Le}]{yang-2019-xlnet}
Zhilin Yang, Zihang Dai, Yiming Yang, Jaime Carbonell, Russ~R. Salakhutdinov,
  and Quoc~V. Le. 2019.
\newblock {XLNet: Generalized Autoregressive Pretraining for Language
  Understanding}.
\newblock In \emph{Proceedings of the Conference on Neural Information
  Processing Systems (NeurIPS)}.

\bibitem[{Zellers et~al.(2018)Zellers, Bisk, Schwartz, and
  Choi}]{zellers-etal-2018-swag}
Rowan Zellers, Yonatan Bisk, Roy Schwartz, and Yejin Choi. 2018.
\newblock {SWAG: A Large-Scale Adversarial Dataset for Grounded Commonsense
  Inference}.
\newblock In \emph{Proceedings of the Conference on Empirical Methods in
  Natural Language Processing (EMNLP)}.

\bibitem[{Zellers et~al.(2019)Zellers, Holtzman, Bisk, Farhadi, and
  Choi}]{zellers-etal-2019-hellaswag}
Rowan Zellers, Ari Holtzman, Yonatan Bisk, Ali Farhadi, and Yejin Choi. 2019.
\newblock {HellaSWAG: Can a Machine Really Finish Your Sentence?}
\newblock In \emph{Proceedings of the Annual Meeting of the Association for
  Computational Linguistics (ACL)}.

\end{thebibliography}
\bibliographystyle{acl_natbib}

\clearpage
\appendix

\section{Dataset Splits}
\label{sec:splits}

Dataset splits are shown in Table \ref{table:splits}.

\begin{table}[h]
\setlength{\tabcolsep}{4pt}
\small
\centering
\begin{tabular}{lrrr}
\toprule
Dataset & Train & Dev & Test \\
\midrule
SNLI & 549,368 & 4,922 & 4,923 \\
MNLI & 392,702 & 4,908 & 4,907 \\
QQP & 363,871 & 20,216 & 20,217 \\
TwitterPPDB & 46,667 & 5,060 & 5,060 \\
SWAG & 73,547 & 10,004 & 10,004 \\
HellaSWAG & 39,905 & 5,021 & 5,021 \\
\bottomrule
\end{tabular}
\caption{Training, development, and test dataset sizes for SNLI \cite{bowman-etal-2015-large}, MNLI \cite{williams-etal-2018-broad}, QQP \cite{iyer-2017-quora}, TwitterPPDB \cite{lan-etal-2017-continuously}, SWAG \cite{zellers-etal-2018-swag}, and HellaSWAG \cite{zellers-etal-2019-hellaswag}.}
\label{table:splits}
\end{table}

\section{Training and Optimization}
\label{sec:training-and-optimization}

For non-pre-trained model baselines, we use the open-source implementations of DA \cite{parikh-etal-2016-decomposable} and ESIM \cite{chen-etal-2017-enhanced} in AllenNLP \cite{gardner-etal-2018-allennlp}, except in the case of SWAG/HellaSWAG, where we run the baselines available in the authors' code.\footnote{\url{https://github.com/rowanz/swagaf}} For BERT \cite{devlin-etal-2019-bert} and RoBERTa \cite{liu2019roberta}, we use \texttt{\small{bert-base-uncased}} and \texttt{\small{roberta-base}}, respectively, from HuggingFace Transformers \cite{wolf2019huggingfaces}. BERT is fine-tuned with a maximum of 3 epochs, batch size of 16, learning rate of 2e-5, gradient clip of 1.0, and no weight decay. Similarly, RoBERTa is fine-tuned with a maximum of 3 epochs, batch size of 32, learning rate of 1e-5, gradient clip of 1.0, and weight decay of 0.1. Both models are optimized with AdamW \cite{loshchilov2018decoupled}. Other than early stopping on the development set, we do not perform additional hyperparameter searches. Finally, all experiments are conducted on NVIDIA V100 32GB GPUs, with the total time for fine-tuning all models being under 24 hours.

Furthermore, temperature scaling line searches are performed in the range [0.01, 5.0] with a granularity of 0.01. These searches are quite fast and can be performed on a CPU; we simply evaluate calibration error by rescaling cached logits. On a Intel Xeon E3-1270 v3 CPU, all searches can be completed in under 15 minutes.

\begin{table}[t]
\setlength{\tabcolsep}{4pt}
\small
\centering
\begin{tabular}{lrrrr}
\toprule
\multirow{2}{*}{Model} & \multicolumn{2}{c}{Accuracy} & \multicolumn{2}{c}{ECE} \\
\cmidrule(lr){2-3} \cmidrule(lr){4-5}
 & ID & OD & ID & OD \\
\midrule
\multicolumn{5}{l}{\textbf{Task: SNLI/MNLI}} \\
\midrule
BERT & 90.18 & 74.04 & 3.43 & 8.18 \\
RoBERTa & \textbf{91.20} & \textbf{79.17} & \textbf{1.18} & \textbf{1.41} \\
\midrule
\multicolumn{5}{l}{\textbf{Task: QQP/TwitterPPDB}} \\
\midrule
BERT & \textbf{90.22} & 86.02 & 4.68 & 11.30 \\
RoBERTa & 89.97 & \textbf{86.17} & \textbf{3.09} & \textbf{9.57} \\
\midrule
\multicolumn{5}{l}{\textbf{Task: SWAG/HellaSWAG}} \\
\midrule
BERT & 78.82 & 38.01 & \textbf{2.51} & \textbf{2.24} \\
RoBERTa & \textbf{81.85} & \textbf{59.03} & 3.02 & 5.71 \\
\bottomrule
\end{tabular}
\caption{Out-of-the-box calibration \textit{development set} results for in-domain (SNLI, QQP, SWAG) and out-of-domain (MNLI, TwitterPPDB, HellaSWAG) datasets using pre-trained models.}
\label{tab:dev-results}
\end{table}

\section{Reproducibility}

Table~\ref{tab:dev-results} shows the accuracy and expected calibration error (ECE) of BERT and RoBERTa on the development sets of the datasets we consider. We do not report post-hoc calibration results using the development set since these require tuning on the development set itself.

\end{document}